\documentclass[twoside,leqno,twocolumn]{article}

\usepackage[letterpaper]{geometry}

\usepackage{ltexpprt}
\usepackage{hyperref}
\usepackage{stmaryrd}
\usepackage{times}
\usepackage{helvet}
\usepackage{courier}
\usepackage{latexsym}
\usepackage{booktabs}
\usepackage{subfigure}
\usepackage{enumitem}
\usepackage{footmisc}
\usepackage{balance}
\usepackage{adjustbox}
\usepackage{amsmath,amssymb}
\usepackage{multirow}
\usepackage{url}

\usepackage{graphicx}

\begin{document}

\newcommand\relatedversion{}
\renewcommand\relatedversion{\thanks{The full version of the paper can be accessed at \protect\url{https://arxiv.org/abs/1902.09310}}} 

\title{\huge Word Embedding with Neural Probabilistic Prior}

\author{\textbf{Shaogang Ren, Dingcheng Li, Ping Li} \\
Cognitive Computing Lab\\
Baidu Research\\
10900 NE 8th St. Bellevue, WA 98004, USA\\
  \texttt{\{renshaogang, dingchengl,  pingli98\}@gmail.com}
}

\date{
}
\maketitle

\begin{abstract}
To improve word representation learning, we propose a probabilistic prior which can be seamlessly integrated with word embedding models. Different from previous methods, word embedding is taken as a probabilistic generative model,  and it enables us to impose a prior regularizing word representation learning. The proposed prior not only enhances the representation of embedding vectors but also improves the model's robustness and stability. The structure of the proposed prior is simple and effective,  and it can be easily implemented and flexibly plugged in most existing word embedding models. Extensive experiments show  the proposed method improves word representation on various tasks.
\end{abstract}


\section{Introduction}

Unsupervised  word or item embedding~\cite{mikolov2013distributed,pennington2014glove,levy2014neural,tang2014learning,gupta2019better}    yields the essential  representation for down-stream information retrieval systems or natural language processing models. Recent progress on pre-trained models significantly improved language processing tasks~\cite{devlin2019bert,yang2019xlnet,radford2019language}. To further improve  word representation learning, we  propose a neural probabilistic prior to integrate   generative models and word representation learning. We first briefly review word embedding  and generative models.

\subsection{Word Embedding}

Classical shallow word embedding methods such as Skip-gram and CBow~\cite{mikolov2013distributed}, and GloVe~\cite{pennington2014glove}  learn word embedding based on the occurrence of words in a sliding window.   These confessional word embedding methods learn word semantics based on co-occurrences and typically cannot  capture the direct structural information in the sequences. The learned word representation can show some simple semantic properties, and they are generally  taken as  the input for down-stream black-box neural network models. Shallow  embedding models may capture  simple semantic information, but they may lose  complicated syntactic and semantic information  embodied in sentences and corpora.

 Pre-trained word embedding and language models can overcome the drawbacks of classical word embedding methods.  Pre-trained word embedding models can effectively integrate the learned prior knowledge and the information from the specific tasks in hand~\cite{peters2018deep,devlin2019bert,yang2019xlnet,radford2019language}. These models usually  are capable of capturing the word token order information among the large number of sentences from a corpus  by leveraging recurrent neural networks~\cite{hochreiter1997long} and/or attention mechanism~\cite{vaswani2017attention}. Training of  pre-trained models  comes with high  costs such as  large training corpora, long computation hours, and financial costs. Those  may also reduce the models' flexibility  in application scenarios, e.g., when  the training corpus or dataset is small~\cite{chen2016learning}.

To achieve effectiveness and flexibility of representation learning, researchers try to incorporate syntactic and semantic information to shallow or small models~\cite{vashishth2019incorporating,faruqui2015retrofitting,marcheggiani2017encoding,mikolov2013linguistic,komninos2016dependency}. These structural information can be flexibly encoded in the learned representation with graph convolutional neural networks~(GCNs)~\cite{kipf2017semi}. In~\cite{vashishth2019incorporating}, they improve word embedding by leveraging both syntactic and semantic  structural information  extracted from the train corpus. Their model achieves superior performance on 3 task categories: word similarity,  word analogy, and word categorization. They claim the method is a CBow model equipped with GCNs. People also try to improve the interpretability of embeddings  by imposing structure regularization on embedding vectors by leveraging explicit word labels~\cite{liao2020explanining}.

\subsection{Variational Autoencoder}

The variational autoencoder~(VAE)~\cite{kingma2014auto} is a  generative model  for unsupervised latent representation learning. Its training algorithm tries to maximize the  following lower bound of data likelihoods,
\begin{align*}
\min_{\phi, \theta}  \mathbb{E}_{p(\mathbf{x})}
& \big[\mathbb{E}_{q_{\phi}(\mathbf{h}|\mathbf{x})} [\log p_{\theta} (\mathbf{x}|\mathbf{h})]
-   \textbf{KL}(q_{\phi}(\mathbf{h}|\mathbf{x}) || p(\mathbf{h}))  \big].
\end{align*}
Here $\mathbf{h}$ is the latent representation of data sample $\mathbf{x}$. The first term of the objective is data reconstruction, and the second one~(\textbf{KL}) is to regularize latent representation $\mathbf{h}$ with a prior $p(\mathbf{h}))$.   People use variants of VAE to obtain disentanglement representation from different data sets including images and texts.  Recently, people utilize nonlinear independent component analysis~(ICA) theory~\cite{hyvarinen2017nonlinear,khemakhem2020variational,halva2020hidden} to improve disentangling representation learning. This line of works is based on the theory that  the latent factors of data distribution can be approximately recovered by leveraging  weak  data labels or data structure information.

\subsection{Contributions}
We propose a novel approach to improve word representation by leveraging statistical  regularization techniques. Different from classical deterministic approaches, we view word embedding as a probabilistic generative model, concretely,  conditional variational autoencoder~(CVAE)~\cite{sohn2015learning}. This perspective allows us to impose a probabilistic prior that enables the model to learn independent latent factors  generating embedding vectors. The proposed representative prior's mean and variance  are parameterized with a neural network.   Alternatively, it can be taken as a  regularization over the learning of word representation based on recent nonlinear ICA theory~\cite{khemakhem2020variational,halva2020hidden,hyvarinen2017nonlinear,sorrenson2020disentanglement}. Our method can be easily implemented and plugged to existing word embedding models. Experiments on various evaluation tasks validate the advantages of the proposed method.

\section{Methodology }\label{sec:cvae}

We first investigate word embedding from conditional VAE perspective. Then we introduce the proposed prior based on conditional VAE and nonlinear ICA theory.

\subsection{Conditional VAE View of Word Embedding}

Most classical word embedding methods~\cite{mikolov2013distributed,pennington2014glove,levy2014neural,tang2014learning,gupta2019better} can be categorized as variants of conditional variational autoencoder~(CVAE). For a word token sequence $\mathbf{s} = [w_1, ..., w_n]$,
let $\tilde{\mathbf{s}}$ be the corrupted sequence~(word tokens with masks) of $\mathbf{s}$ with $\mathbf{m}$ as the corresponding binary masks on words. Here we take $\mathbf{y}=\{\mathbf{s}, \mathbf{m}\}$ as the label information regarding $\tilde{\mathbf{s}}$, and try to model the conditional distribution $p(\tilde{\mathbf{s}}|\mathbf{y})$ with conditional variational autoencoder.  The objective of CVAE~is:
\begin{align}\notag
&\mathcal{L}_{\text{CVAE}}(\tilde{\mathbf{s}},\mathbf{y}; \theta, \psi, \phi) 
= \mathbb{E}_{q_{\phi}(\mathbf{h}|\tilde{\mathbf{s}}, \mathbf{y})} \big[\log p_{\theta}(\tilde{\mathbf{s}}|\mathbf{h}, \mathbf{y}) \big]  \\ \notag
&\hspace{0.9in}-\textbf{KL}(q_{\phi}(\mathbf{h}|\tilde{\mathbf{s}},\mathbf{y}))|| p_{\psi}(\mathbf{h}|\mathbf{y})
\leq  p_{\phi, \psi}(\tilde{\mathbf{s}}|\mathbf{y}) .
\end{align}
For conventional word embedding methods, $\mathbf{h}$ is the sequential concatenation of  unmasked words' embedding vectors. The encoder~($q_{\phi}$) is an indicator function converting unmasked tokens to their embedding vectors. The decoder~($p_{\theta}$) maps from  $\mathbf{h}$ to the~masked tokens.

\begin{figure}[h]
\begin{center}
  \includegraphics[width=2.9in]{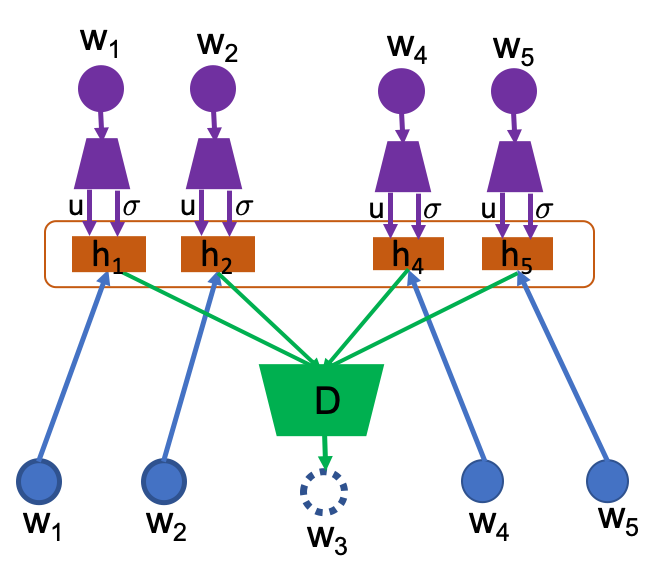}
\end{center}

\vspace{-0.3in}

  \caption{ \textbf{Word embedding with neural probabilistic prior.} $w_3$ is the masked word, $w_1$,  $w_2$, $w_4$, and $w_5$ are context words. Blue, green, and purple lines represent, encoding,  decoding, and  prior information, respectively. }
\label{fig:dwe}\vspace{-0.15in}
\end{figure}

As shown in Figure~\ref{fig:dwe}, the blue lines are  the encoder, $q_{\phi}$, and green lines are the decoder,  $p_{\theta}$. In classical embedding methods,
both encoder and decoder are deterministic functions. There is no prior~($p_{\psi}$) for word embedding~$\mathbf{h}$ in conventional methods, and thus the $\textbf{KL}$ term is vanished.

In this paper, we will add  the prior~($p_{\psi}$) of $\mathbf{h}$ with a real distribution function~(neural network) and impose regularization on embedding vectors. The prior enables the model to learn latent independent factors across the embedding entries. The decoder can be implemented with a neural network or graph convolutional network~(GCN)~\cite{kipf2017semi}. By leveraging syntactic and semantic structure information, GCNs can learn task agnostic word representations. GCNs are flexible to aggregate structural knowledge among the words, and this improves the  representation of learned word embedding vectors. The proposed model takes work tokens as word embedding labels, and it tries to naturally  learn the independent factors without  explicit labeling or structure constrain on the embedding vectors. Thus our method is significantly different from the approach in~\cite{liao2020explanining} that embedding vectors are manually segmented into different sections, and each section represent one attribute of the words.  Moreover, their method also requires explicit labels for word attributes that can be ignored in our model.

\subsection{Neural Probabilistic Prior}

We introduce a  word embedding distribution prior~(WEP) to the   CVAE model, and the prior is parameterized with a neural network.

\subsubsection{Probabilistic Prior with Nonlinear ICA}

 On observation of the conditional VAE framework of word embedding, we follow the recently developed nonlinear ICA theory~\cite{khemakhem2020variational,halva2020hidden,hyvarinen2017nonlinear} to improve word embedding. People have implemented this type of prior to improve the disentanglement and representation of images~\cite{sorrenson2020disentanglement}.  By leveraging  weak  data labels or data structure information, a nonlinear ICA prior learns the independent factors generating data latent variables, and hence it  improves the disentanglement of data representation. In this paper, word embedding vectors are taken as the hidden representation and are expected to be improved by leveraging the sequential label information $\mathbf{y}=\{\mathbf{s}, \mathbf{m}\}$.

For word $w_t$, suppose the distribution regarding $\mathbf{h}_t$ is a factorial member of the exponential family with $v$ sufficient statistics, conditioned on word token $w_t$. The general form of $\mathbf{h}_t$ prior distribution $p_{\psi}$ can be written~as
\begin{align}\notag
&p_{\psi}(\mathbf{h}_t | w_t)
=\Pi_{i=1}^l \frac{Q_i(h_{t,i})}{Z_i(w_t)} \text{exp}\bigg[ \sum_{j=1}^v T_{i,j}(h_{t,i}) \lambda_{i,j}(w_t) \bigg].
\end{align}
Here $Q_i$ is the base measure, $Z_i$ is the normalizing constant, $T_{i,j}$s are the sufficient statistics and $\lambda_{i,j}$ the corresponding parameters,  depending on $w_t$.
Let $\tilde{\mathbf{s}}$ be the output of an arbitrarily complex, inevitable, and deterministic decoder from the latent space to the masked word tokens or missing tokens, i.e., $\tilde{\mathbf{s}} = D(\mathbf{h})$.  Using nonlinear ICA theory~\cite{khemakhem2020variational}, with enough training samples, the conditional VAE can linearly uncover the  factors generating the word embedding vector $\mathbf{h}_t$,~i.e.,
\begin{align*}
&\mathbf{T}(\mathbf{h}_t) = \mathbf{A}\widehat{\mathbf{T}}(\widehat{\mathbf{h}}_t) + \mathbf{c}, \quad \mathbf{Q} = \mathbf{A}\widehat{\mathbf{Q}}.
\end{align*}
Here $\widehat{\mathbf{T}}$, $\widehat{\mathbf{h}}_t$, and $\widehat{\mathbf{Q}}$ are learned from the model, and they correspond to ground truth $\mathbf{T}$, $\mathbf{h}_t$, and $\mathbf{Q}$,  respectively. $\mathbf{A}$ is a full rank matrix and $ \mathbf{c}$ is a constant vector. Hence the base measure $\mathbf{Q}$ and sufficient statistics $\mathbf{T}$ regarding the word embedding vector $\mathbf{h}_t$ can be linearly recovered by the model as long as there are enough word token labels regarding the conditional prior distribution $p_{\psi}(\mathbf{h}_t | w_t)$.

This paper proposes an approach to impose the prior distribution on  word embedding via above nonlinear ICA  framework. We employ Gaussian distribution as the prior $p_{\psi}(\mathbf{h}_t | w_t)$, and the mean $\mu$ and variance $\sigma^2$ are parameterized by a neural network with  word token~($w_t$) as the input~(all words use the same neural network for the prior). Neural network $p_{\psi}$ accumulates knowledge across all the words and can linearly identify independent latent factors  generating the vocabulary.

\subsubsection{Implementation}

We take word embedding models as conditional VAEs, and the context words~(tokens) as the label regarding a sequence of masked sentence.  The concatenation of context words' embedding vectors~($\mathbf{h}$) is the latent variable of the conditional VAE model. Figure~\ref{fig:dwe} shows a diagram of the proposed method. We add an exponential family prior to each word's embedding vector, and the prior is dependent on the word's label information. As discussed previously, this paper  chooses Gaussian as the prior distribution for embedding~vectors. We start from a sequence context model~(sliding window) and then extend it to graph context cases.

Similar to the continuous-bag-of-words (CBOW) model~\cite{mikolov2013linguistic}, we define the context of a word $w_k$ as $\mathbf{c}_{w_k} =\{w_{k+j}: -c \leq j \leq c, j \neq 0 \}$ for a window with size c. For a word $w_t \in \mathbf{c}_{w_k}$, the prior distribution parameters regarding embedding vector $\mathbf{h}_t$ are $\mathbf{\mu}(w_t)$ and $\mathbf{\sigma}(w_t)$.
$\mathbf{\mu}(w_t)$ and $\mathbf{\sigma}(w_t)$ are the mean and standard deviation,  respectively, and they are parameterized with a neural network with word token $w_t$ as the input.  The prior distribution can be integrated with word embedding under the conditional VAE framework discussed in previous sub-sections.
The objective regarding a draw of $w_k$ and  $\mathbf{c}_{w_k}$ is written~as
\begin{align*}
& \mathcal{L}(w_k, \mathbf{c}_{w_k}; \theta, \psi, \mathbf{h} )
=\mathbb{E}_{q(\mathbf{h}|\mathbf{c}_{w_k})} \big[\log p_{\theta}(w_k|\mathbf{h}) \big] \\ \notag
&\quad \quad - \textbf{KL}\big(q(\mathbf{h}|\mathbf{c}_{w_k})|| p_{\psi}(\mathbf{h}|\mathbf{c}_{w_k})\big) .
\end{align*}
$q(\mathbf{h}|\mathbf{c}_{w_k})$ is a deterministic indicator function. The \textbf{KL} can be replaced with Gaussian likelihood:
\begin{align}  \notag
& \mathcal{L}(w_k, \mathbf{c}_{w_k}; \theta, \psi, \mathbf{h} )
=\mathbb{E}_{q(\mathbf{h}|\mathbf{c}_{w_k})} \big[\log p_{\theta}(w_k|\mathbf{h}) \big]  \\ \label{eq:obj}
&\hspace{0.2in} -\alpha \sum_{w_i \in \mathbf{c}_{w_k}} \bigg\{ \frac{1}{2} \bigg( \frac{ \mathbf{h}_i - \mathbf{\mu}(w_i)}{\sigma(w_i)}\bigg)^2 + \log \mathbf{\sigma}(w_i) \bigg\},
\end{align}
where $\alpha$ is a non-negative value controlling the impact of the prior. The reconstruction term $\mathbb{E}_{q(\mathbf{h}|\mathbf{c}_{w_k})} \big[\log p_{\theta}(w_k|\mathbf{h}) \big]$ can be implemented with any word embedding model's loss function.  The distribution parameter neural networks $\mathbf{\mu}$ and $\mathbf{\sigma}$ statistically aggregate information across all the words' structure distributions, and uncover  later factors that generate word representations.  The neural network structure for the prior $p_{\psi}$ is presented in Table~\ref{tab:prior}.
Word token ID is an integer, and they are duplicated multiple times  to form an input vector for the probabilistic prior neural network.  The  output of prior network is evenly divided into two parts, $\mathbf{\mu}$ and $\log \mathbf{\sigma}$. The  objective~\eqref{eq:obj} has two sets of parameters,  $\theta$ and $\psi$, and they correspond to the parameter of the decoder and the prior. In this paper, we employ SynGCN and SemGCN in~\cite{vashishth2019incorporating} for the reconstruction term in~\eqref{eq:obj}. In both models, a GCN  is employed for the decoder in order to incorporate the syntactic and semantic structural knowledge.

\begin{table}[h]

\vspace{-0.1in}
\begin{center}
\caption{\label{tab:prior} \textbf{Network structure for our neural probabilistic prior.} FC stands for fully-connected layer.\vspace{0.05in}}
\begin{tabular}{ l  | c | c | c  }\hline
\multicolumn{4}{ c }{The prior function $p_{\psi}$  } \\
\hline
Layer  & Dimension of & Batch & Activation  \\
 & Output  &    Normalization & function \\
\hline
Input $w_t$ &  $32$ & - & - \\
\hline
FC & 64 & N   & ReLU \\
\hline
FC & 2$\times$dim($\mathbf{h}$) & N &  -  \\
\hline
\end{tabular}
\end{center}\vspace{-0.1in}
\end{table}

The proposed method uses SynGCN and SemGCN~\cite{vashishth2019incorporating} as the base models. Word embedding can be improved through incorporating  graphical syntactic and semantic information, and the context words' sliding window is replaced by or combined  with neighbors in the context syntactic or semantic graph.  For a given sentence $\mathbf{s}=[w_1, w_2, ..., w_n]$, we first extract its dependency parse graph $\mathcal{G}_{\mathbf{s}} = (\mathcal{V}_{\mathbf{s}}, \mathcal{E}_{\mathbf{s}} )$ using  Stanford CoreNLP parser. Here, $\mathcal{V}_{\mathbf{s}} = \{w_1,w_2,...,w_n\}$ and $\mathcal{E}_{\mathbf{s}}$ denotes the labeled directed dependency edges of the form $(w_i,w_j,l_{ij})$, where $l_{ij}$ is the dependency relation of $w_i$ to $w_j$. We define the context of a word $w_k$ as its neighbors in $\mathcal{G}_{\mathbf{s}}$, i.e., $\mathbf{c}_{w_k} = \mathcal{N}(w_k)$. Besides syntactic information, word embedding can be further improved  by incorporating semantic knowledge.  The decoder learns a corpus-level  labeled graph with words as nodes and edges representing semantic relationship among them from different sources. Following SemGCN~\cite{vashishth2019incorporating}, semantic relations such as hyponymy, hypernymy and synonymy are represented together in a single graph.  We use {\it WEPSyn} and {\it WEPSem} to represent the proposed models incorporating with syntactic and semantic information, respectively. WEPSyn and WEPSem  use the same objective~\eqref{eq:obj} and the same prior network in Table~\ref{tab:prior}.

\section{Experiments}

We compare  the proposed method against existing methods on standard datasets. The prior network~($\mathbf{u}$ and $\mathbf{\sigma}$) for both  WEPSyn and  WEPSem are given in Table~\ref{tab:prior}. We  vary $\alpha$ in~\eqref{eq:obj} to tune the regularization of the prior distribution. In our experiments, we use $\alpha \in \{0.5, 0.1,  1.0e\text{-}4, 1.0e\text{-}6\}$.

 WEPSyn can be taken as the SynGCN~\cite{vashishth2019incorporating} model plugged with the proposed prior, and similarly WEPSem is the SemGCN~\cite{vashishth2019incorporating} model enhanced with the  prior. We compare WEPSyn and WEPSem with SynGCN and SemGCN methods that have no prior regularization in the experiments to validate the advantages of the proposed prior.

\subsection{Evaluation Methods}\label{sec:eval}

We consider the following baseline methods to compare with WEPSyn:
\begin{itemize}

 \item \textbf{Word2vec}~\cite{mikolov2013linguistic} is continuous-bag-of-words.

 \item \textbf{GloVe}~\cite{pennington2014glove}, a log-bilinear regression model which leverages global co-occurrence statistics of corpus.

\item  \textbf{Deps}~\cite{levy2014dependency} is a modification of skip-gram model which uses dependency context in place of sequential context.
\item \textbf{EXT}\cite{komninos2016dependency} is an extension of Deps which utilizes second-order dependency context features.
\item \textbf{SynGCN}\cite{vashishth2019incorporating} leverages GCN and syntactic word relationships to improve word embedding.
\end{itemize}
WEPSem  model is evaluated against the following methods:
\begin{itemize}

\item \textbf{Retro-fit}~\cite{faruqui2015retrofitting} is a post-processing procedure using similarity constraints from semantic knowledge sources.

\item \textbf{Counter-fit}~\cite{mrksic2016counter} is a method for injecting both antonym and synonym constraints into word embeddings.

\item \textbf{JointReps}~\cite{alsuhaibani2018jointly} is a joint word representation learning method by simultaneously utilizing the corpus~and~KB.

\item \textbf{SemGCN}~\cite{vashishth2019incorporating} leverages GCN and semantic word relationships to improve word embedding.

\end{itemize}

\subsubsection{Intrinsic Tasks}\label{sec:intrinsic_task}

\begin{itemize}
\item \textbf{Word Similarity} is the task of evaluating closeness between semantically similar words. Following~\cite{vashishth2019incorporating,komninos2016dependency,pennington2014glove},  we perform comparison of different methods on SimLex999~\cite{hill2015simlex}, WS353~\cite{finkelstein2001placing}, and RW~\cite{luong2013better} datasets.

\item \textbf{Word Analogy} task is to predict word $b_2$, given three words $a_1$, $a_2$, and $b_1$, such that the relation b1 : b2 is same as the relation a1 : a2. We compare our method to the baselines  on  SemEval2012~\cite{jurgens2012semeval} and MSR~\cite{mikolov2013linguistic}.

\item \textbf{Concept Categorization} involves grouping nominal concepts into natural categories. We evaluate on AP~\cite{almuhareb2006attributes}, Battig~\cite{baroni2010distributional}, BLESS~\cite{baroni2011how}, ESSLI~\cite{baroni2008esslli} datasets.
\end{itemize}

\subsubsection{Extrinsic Tasks}\label{sec:extrinsic_task}

\begin{itemize}
\item \textbf{Named Entity Recognition (NER)} is the task of locating and classifying entity mentions into categories like person, organization etc.
We use~\cite{lee2018higher}'s model on CoNLL-2003 dataset~\cite{sang2003introduction}  for evaluation.

\item \textbf{Question Answering}  in Stanford Question Answering Dataset (SQuAD)~\cite{rajpurkar2016squad} involves identifying answer to a question as a segment of text from a given passage. Following~\cite{peters2018deep}, we evaluate using \cite{clark2018simple}'s model.

\item \textbf{Part-of-speech (POS) tagging} aims at associating with each word, a unique tag describing its syntactic role. For evaluating  embeddings, we use ~\cite{lee2018higher}’s model on Penn Treebank POS dataset~\cite{marcus1994penn}.

\item \textbf{Co-reference Resolution (Coref)} involves identifying all expressions that refer to the same entity in the text. To inspect the effect of embeddings, we use \cite{lee2018higher}’s model on CoNLL-2012 shared task dataset~\cite{pradhan2012conll}.
\end{itemize}

\subsection{Results}

We first present  results on comparison WEPSyn against the baselines. We then show the results to compare  WEPSem and SemGCN.

\begin{table*}[h!]
\begin{center}
\caption{\textbf{WEPSyn intrinsic evaluation.} Performance on word similarity (Spearman correlation), concept categorization (cluster purity), and word analogy (Spearman correlation). Our   WEPSyn outperforms existing methods on 8 out 10  settings.\vspace{0.05in}}\label{tab:res}
{\small
\begin{tabular}{c  c  c  c  c | c  c  c  c | c  c }\hline
 & \multicolumn{4}{c}{\textbf{Word Similarity}} & \multicolumn{4}{c}{\textbf{Concept Categorization}}  &\multicolumn{2}{c}{\textbf{Word Analogy}} \\
  \cline{2-5}  \cline{6-9}  \cline{10-11}
  Method & WS353S & WS353R & SimLex999 & RW & AP & Battig & BLESS &  ESSLI & SemEval2012  & MSR \\
Word2vec &71.4 & 52.6 &38.0 & 30.0 &  63.2 & 43.3 &77.8 & 63.0 & 18.9 & 44.0 \\
GloVe &69.2 & \textbf{53.4}   &36.7  & 29.6  &58.0  &41.3    &80.0 &59.3. &18.7  &45.8   \\
Deps & 65.7  &36.2   &39.6  & 33.0  & 61.8  & 41.7   &65.9 & 55.6  & 22.9  &40.3 \\
EXT & 69.6  & 44.9   &43.2  &18.6   &52.6  & 35.0  & 65.2  & 66.7 & 21.8  &18.8 \\
SynGCN &73.2 & 45.7 &45.5 & 33.7 &  69.3  & 45.2 & 85.2 & 70.4 &  23.4 &  \textbf{52.8} \\
\hline
WEPSyn & \textbf{75.7} &  47.7 &  \textbf{46.9} &   \textbf{36.0} &   \textbf{73.8} &  \textbf{46.9} &  \textbf{86.0} &  \textbf{79.7} &  \textbf{23.8} & 46.4 \\
\hline
\end{tabular}
}
\end{center}\vspace{-0.15in}
\end{table*}

\subsubsection{WEPSyn}\label{sec:wepsyn}

In this set of experiments, WEPSyn and SynGCN  use the same syntactic knowledge extracted from corpus.
Following~\cite{vashishth2019incorporating,mikolov2013linguistic,komninos2016dependency}, we separately define target and context embeddings for each word in the vocabulary.  After preprocessing, the Wikipedia corpus consists of 57 million sentences with 1.1 billion tokens and 1 billion syntactic dependencies, and the average sentence length is about 20.

\textbf{WEPSyn Intrinsic Evaluation.} Table~\ref{tab:res} gives the performance of different~methods. From Table~\ref{tab:res}, we can see that the proposed word embedding method ~(WEPSyn)  performs better than existing methods on 8 out 10 tasks.  WEPSyn achieves the largest improvement on 4 concept categorization tasks compared with all the baseline methods.~It~means~the proposed prior significantly augments the model's ability to capture concept structure. WEPSyn performs better than the baselines on three word similarity tasks, WS353S, SimLex999, and RW. Particularly,  WEPSyn has higher scores on all four word similarity tasks in comparison with SynGCN, which also utilizes word syntactic information. WEPSyn obtains at least 1.4\% and 0.8\%  performance increase on word similarity and concept categorization with the help of regularization prior.
For word analogy task, WEPSyn can achieve comparable results  with other methods. As mentioned above, WEPSyn can be taken as the SynGCN model plugged with the proposed prior, and hence the experimental results in Table~\ref{tab:res} indicate that the proposed prior helps the model to identify  independent latent factors of word embedding, and hence the model achieves better results on most settings in comparison with SynGCN~\cite{vashishth2019incorporating}, especially the concept categorization tasks. Without a prior to regularize word representation learning, existing methods, including SynGCN, cannot outperform our method in most tasks in the experiments.

\textbf{WEPSyn Extrinsic Evaluation.}  We further investigate  WEPSyn on downstream tasks~(extrinsic tasks) defined in section~\ref{sec:extrinsic_task}. Table~\ref{tab:wepsyn_extrinsic} shows the comparison  between WEPSyn and the existing baselines on the four extrinsic tasks.
We find that WEPSyn(SG) achieves comparable or  better results than the baselines on all tasks. The proposed neural probabilistic prior helps the embedding model to accumulate the knowledge  of all words across  different
sequences and word usages, and it improves the  representation of words and hence the performances on different tasks.

\begin{table}[h!]
\begin{center}

\caption{\label{tab:wepsyn_extrinsic}\textbf{ WEPSyn extrinsic evaluation.} WEPSyn performs comparable to or outperforms baseline methods on all tasks including  parts-of-speech tagging (POS), question answering (SQuAD), named entity recognition (NER), and co-reference resolution (Coref).  }\vspace{0.05in}
{
\begin{tabular}{ l  c  c   c  c }\hline
Method  & POS  & SQuAD & NER & Coref \\
\hline
Word2vec & 95.0 &78.5& 89.0 & 65.1 \\
GloVe &94.6&78.2& 89.1 &64.9 \\
Deps & 95.0 &77.8&88.6 &64.8 \\
EXT & 94.9 &79.6&88.0 &64.8 \\
SynGCN &95.4  & 79.6 & 89.5 & 65.8 \\
\hline
WEPSyn &\textbf{95.5} & \textbf{80.1} & \textbf{90.0} & \textbf{66.2}\\
\hline
\end{tabular}}
\end{center}\vspace{-0.15in}
\end{table}

\subsubsection{WEPSem}

 In this set of experiments, WEPSem and the baselines use  semantic information and are initialized with  pre-trained embeddings to improve word embedding. As mentioned above, the training of WEPSem is also regularized with the proposed neural probabilistic prior given in objective~\eqref{eq:obj} with  the neural network structure in Table~\ref{tab:prior}.
Following~\cite{vashishth2019incorporating}, the \emph{hypernym, hyponym}, and \emph{antonym} relations from WordNet, and \emph{synonym} relations from PPDB are used in the experiments.

Each method is enhanced with  the semantic information it  can make use of. For example, Retro-fit can only  use the synonym relation from PPDB. Here $M(I, S)$ represents  the fine-tuned embeddings learned with  model $M$ while taking $I$ as the initialization embeddings.  The type of semantic information used is denoted by $S$. Following~\cite{vashishth2019incorporating}, wet let:
\begin{itemize}
\item $S=1$ Only synonym information.

\item $S=2$ Synonym and antonym information.

\item $S=4$ Synonym, antonym, hypernym and hyponym information.
\end{itemize}
For example, Counter-fit(Word2Vec,2) stands for  Word2Vec embedding is fine-tuned by Counter-fit using synonym and antonym information.

\begin{table}[h]
\begin{center}
\caption{\label{tab:wepsem_instrinsic} \textbf{WEPSem intrinsic evaluation.} Evaluation of different methods for incorporating diverse semantic constraints initialized using pre-trained embeddings. WEPSem~(with SynGCN or WEPSyn) outperforms the other methods on all the  tasks. Pre-trained embedding: SG=SynGCN, WS=WEPSyn.\vspace{0.05in}}
{
\begin{tabular}{ l c  c   c  }\hline
Method  & WS353  &AP & MSR \\
\hline
Retro-fit(SG,1) & 61.2 & 67.1 &51.4\\
Counter-fit(SG,2) & 55.2 & 66.4 & 31.7 \\
JointReps(SG,4) & 60.9 & 68.2 & 24.9 \\
SemGCN(SG,4) & 62.1 & 69.3 & 54.4 \\
\hline
WEPSem(SG,4) & 62.6  & 69.7 & 53.7 \\
WEPSem(WS,4)& \textbf{62.9}   &  \textbf{73.8}  & \textbf{55.1} \\
\hline
\end{tabular}}
\end{center}\vspace{-0.15in}
\end{table}

\textbf{WEPSem Intrinsic Evaluation.}
All models are trained using the initial embedding  from SynGCN or WEPSyn.   We use SG and WS to represent the initial word embedding from SynGCN and WEPSyn, respectively.
Table~\ref{tab:wepsem_instrinsic} gives results from four baseline methods and WEPSem on three  tasks, WS353, AP, and MSR. They represent three different evaluation metrics discussed in Section~\ref{sec:eval}.  From Table~\ref{tab:wepsem_instrinsic}, WEPSem(SG,4) achieves comparable performances on three tasks compared with SemGCN(SG,4).  However, WEPSem(WS,4) achieves significantly improved performance on all three tasks compared with SemGCN(SG,4). In practice, the baseline methods barely improves the WEPSyn embedding even with the help of  semantic relations, and this type of results are neglected~here.

With the regularization prior,  WEPSem(WS,4) obtains about 2.0\%  average boost in performance compared with SemGCN(SG,4). WEPSem is equal to a SemGCN model enhanced with the proposed prior.  It means that by leveraging syntactic and semantic information, the proposed prior enables the model to capture independent latent factors to identify word similarity, concept categorization, and word analogy. With more  generalization capacity, WEPSem can improves both SynGCN and WEPSyn  embeddings.

\textbf{WEPSem Extrinsic Evaluation.}
Following extrinsic tasks for WEPSyn in Section~\ref{sec:wepsyn}, we also evaluate WEPSem on extrinsic tasks.
To compare performance on the extrinsic tasks, we first fine-tune SynGCN embeddings using different methods to incorporate semantic information. The embeddings obtained by this process are then evaluated on extrinsic tasks  in Section~\ref{sec:extrinsic_task}. See the results  in Table~\ref{tab:wepsem_extrinsic}.
\begin{table}[h]
\begin{center}

\caption{\label{tab:wepsem_extrinsic} \textbf{WEPSem extrinsic evaluation.} Comparison of different methods for incorporating diverse semantic constraints initialized using pre-trained embeddings on  extrinsic tasks. WEPSem (with  SynGCN or WEPSyn) performs comparable or outperforms all existing approaches on all tasks. Pre-trained embedding: SG=SynGCN, WS=WEPSyn.\vspace{0.05in}}
{
\begin{tabular}{ l c  c   c  c }\hline
Method  & POS  & SQuAD & NER & Coref \\
\hline
SG  &95.4  & 79.6 & 89.5 & 65.8 \\
Retro-fit(SG,1) & 94.8 & 79.6 &88.8 &66.0 \\
Counter-fit(SG,2) &94.7 &79.8 &88.3 &65.7 \\
JointReps(SG,4) & 95.4 &79.4 &89.1 & 65.6 \\
SemGCN(SG,4) &95.5 & 80.4 & 89.5 & 66.1 \\
\hline
WEPSem(SG,4) & 95.5 & 80.5 & \textbf{89.9} & 66.3 \\
WEPSem(WS,4) & \textbf{95.9} & \textbf{80.8} & \textbf{89.9} & \textbf{66.7}\\
\hline
\end{tabular}
}
\vspace{-0.15in}
\end{center}

\end{table}

We can see that WEPSem(SG,4) achieves comparable or better results on all four down stream tasks compared against  SemGCN(SG,4). Different from other baselines, WEPSem(SG,4) can constantly improve SynGCN embeddings on all tasks. We also include WEPSem(WS,4) in Table~\ref{tab:wepsem_extrinsic}, and WEPSem(WS,4)  obtain the best performance on all tasks. These results indicate that the proposed prior indeed helps the backbone model  to learn better representations of the words.

\subsubsection{Comparison with BERT}
We extract a word embedding dictionary from BERT model~\cite{devlin2019bert}  and then we use it as the initial embedding for our WEPSem model. For each word token, its embedding vector is the average of the vectors in different sentence contexts. The results in Table~\ref{tab:bert} compare the BERT and  WEPSem+BERT on four different tasks. We can see that the proposed WEPSem indeed  improves  BERT embedding on different tasks. Please note that only the extracted embedding dictionary (no fine-tuning) of BERT is  used in these tasks.

\begin{table}[h]
\begin{center}

\vspace{-0.1in}

\caption{\label{tab:bert} \textbf{Comparison of WEPSem  with BERT on multiple extrinsic tasks.}  Our proposed method encodes complementary information which is not captured by BERT.\vspace{0.05in} }

\begin{tabular}{ l c  c   c  c }\hline
Method  & POS  & SQuAD & NER & Coref \\
\hline
BERT(B) &95.09  & 79.83 & 89.52 & 66.23 \\
WEPSem(B,4) & \textbf{95.24} & \textbf{80.12} & \textbf{89.90} & \textbf{66.62}\\
\hline
\end{tabular}
\end{center}\vspace{-0.2in}
\end{table}

\subsection{Stabilizing Training with Neural Probabilistic Prior}

\begin{figure}[b!]

\vspace{-0.2in}

\begin{center}
  \includegraphics[width=3.3in]{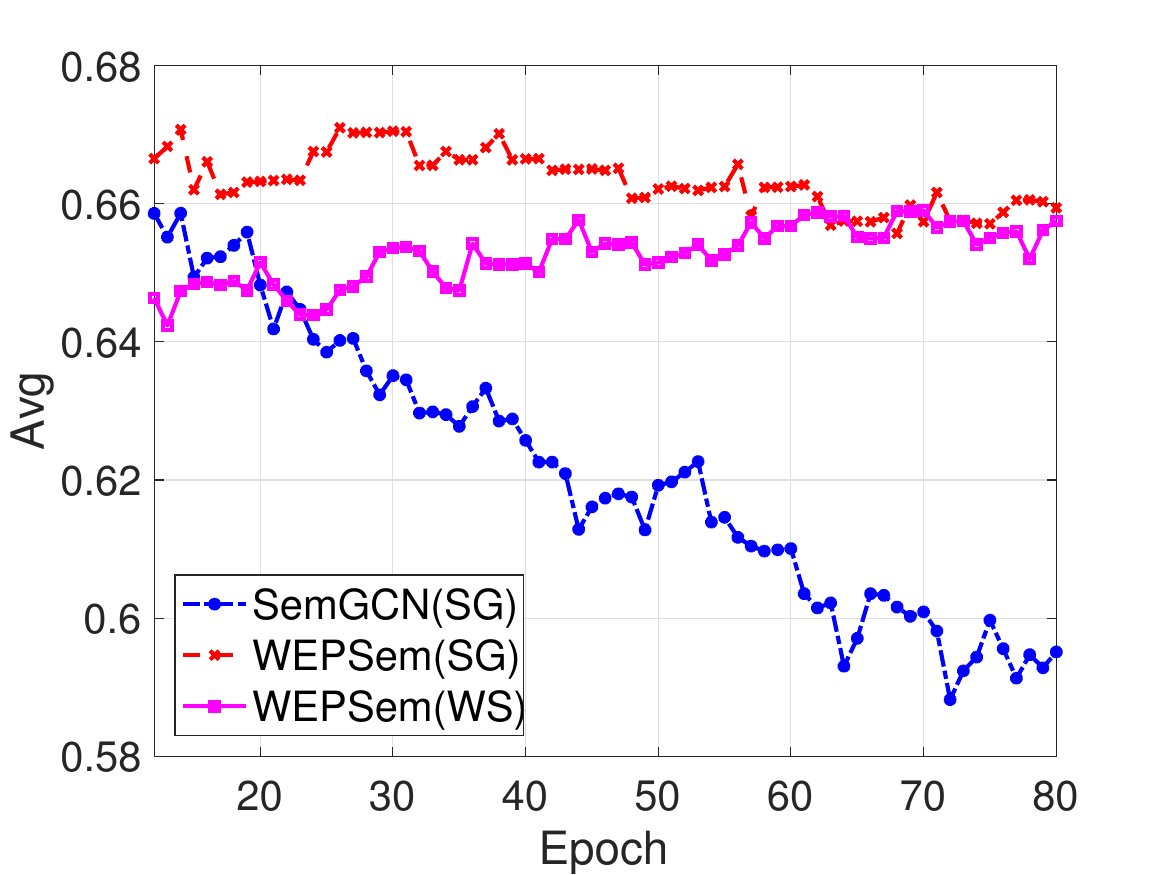}
\end{center}

\vspace{-0.2in}

  \caption{\textbf{Average scores on 3 tasks: WS353,~AP,~MSR  w.r.t. epochs for three  models.}}
\label{fig:epoch}
\end{figure}

\begin{table*}[t]

\begin{center}

\caption{\textbf{Interpretability of word embedding entries.} 8 randomly selected  dimensions are presented. \vspace{0.05in}}\label{tab:interpret_dwe}

{\small
\begin{tabular}{ l | l | l | l }\hline
\textbf{Dim.}  & \textbf{Value Range} & \textbf{Topics} & \textbf{Sample Words} \\
\hline
 210 & [-1.6, -1.4] & Organ & neural, chromosomal, inhibitor, stimulant, pathway, neuronal, oral, cerebral \\
\hline
279 & [-1.5, -1.4] & Life & living, live, plantation, emigrated, boundary, motherland, emigrating, relocation \\
\hline
186 & [-1.4, -1.2] & Farming & irrigate, harvesting, greenland, importation, seas, plants, seabird, basins \\
\hline
24 & [-1.5, -1.3] & Chemistry & oxide, sulphide, chloride, formula, sterile, oxides, sulfide, metallic \\
\hline
 155 &  [-1.0, -0.8] &  Nature & rainforest, jungles, amazon, rainforests, floods, heathland,   extinct, forests \\
\hline
 62 &   [-1.2,-1.1] &  Explosion & explosion, breathing, breathe, suffocation, holocaust,  fighting, \\
 & & & explosions, smothered\\
\hline
 188 &  [1.2,\ \  \    1.3] &  Writing & alphabetical, draft, brushstrokes, diction, poem, expressing,  \\
 & & & textbooks, redrawing  \\
\hline
 233 & [1.4,\ \ \     1.5] & Shape &   hyperbolic, symmetric, transformations, gauss, symmetrical, \\
 & & &  vertices, planes, heart-shaped \\
\hline
\end{tabular}
}
\vspace{-0.2in}
\end{center}
\end{table*}

The proposed prior not only  augments the  learned word embedding, but also  improves the model's stability.  Figure~\ref{fig:epoch} shows the average score of the three tasks, i.e., WS353, AP, and MSR, at different epochs. The performance of WEPSem(WS) continues to increase with the value of training epochs. Without a  prior to regularize the evolving of embedding vectors, SemGCN(SG)'s performance decreases, potentially due to model over-fitting.  Figure~\ref{fig:epoch} indicates that the proposed prior indeed  regularizes the dynamic of embedding vectors and improves model's training stability.

\subsection{Word Embedding Disentanglement Showcase}

We demonstrate disentanglement in the learned  WEPSyn word embeddings. Here, we provide examples to show that for each dimension of word embedding, words with similar entry values automatically group together. The entire value range for each dimension is from about -2 to 2. To find the patterns, we sort the words  according to the increasing order of  each randomly selected embedding dimension. We then query each selected dimension by some words. As shown in Table~\ref{tab:interpret_dwe}, for dimension 210, we queried the word $neural$ and retrieve words nearby and found that quite many nearby words are related to $neural$ in meaning. Similarly, for some ranges of dimension 279, words related to $life$ go together; for dimension 186, words related to farming cluster together; for dimension 24, words related to chemistry form a group.

It indicates that each dimension has certain specific focuses within some value ranges. The results  show that a word in some dimension with different attributes can help discover semantics implicitly encoded dimension by dimension. Similar to disentanglement in images~\cite{sorrenson2020disentanglement},  the proposed neural probabilistic prior helps to disentangle the latent factors in word embeddings  according to the nonlinear ICA theory~\cite{khemakhem2020variational,halva2020hidden,hyvarinen2017nonlinear}. It also implies that the  CVAE perspective of word embedding  in this paper is reasonable.

\section{Comparison with Other  Works}

Our method significantly differs from disentangled word embedding model~\cite{liao2020explanining}. Their method explicitly requires word labels as the supervision to achieve  discrimination between manually separated embedding sections.  Our  prior, however, does not require labeling for disentanglement and embedding sections, can work even with simple embedding methods such as Word2vec~\cite{mikolov2013linguistic}. It takes word tokens as the labels and relies on statistical property of the prior to discover latent structures and  factors of embedding vectors.

Our method also differs from the   Bayesian Skip-gram model~\cite{brazinskas2018embedding}.  The word embedding prior $p(\mathbf{h}_i|w_i)$ for each word $i$  in~\cite{brazinskas2018embedding} is a  Gaussian distribution $\mathcal{N}(\mathbf{h}_i | \mathbf{\mu}_i, \mathbf{\Sigma}_i)$ with covariance $\mathbf{\Sigma}_i$  a simple diagonal matrix. Without the prior neural networks~(structure in Table~\ref{tab:prior}) taking word token ID $i$ as the input  proposed in this paper,  the prior parameters $\{\mathbf{\mu}_i, \mathbf{\Sigma}_i\}$ in~\cite{brazinskas2018embedding} cannot accumulate information across all the word tokens and thus may have restricted  disentangling effect according to nonlinear ICA theory~\cite{khemakhem2020variational,hyvarinen2017nonlinear}. Moreover, our proposed prior is imposed on context words $w_i \in \mathbf{c}_{w_k}$ rather than the central word implemented in~\cite{brazinskas2018embedding}.

In our experiments the proposed prior employs SynGCN and SemGCN in~\cite{vashishth2019incorporating} as the backbone model. The proposed plug-and-play prior can be integrated with any item or word embedding methods for improving the performance. It can be integrated with classical shallow  embedding methods such as Skip-gram and CBow to learn robust representation on small  data. It also can be applied to deep pre-trained or context based learning models  for further improvement.

\section{Conclusion}

This paper presents a new approach for regularizing word representation learning called WEPSyn and WEPSem. The approach combines probabilistic generative models and nonlinear ICA theory and can be applied to various word embedding models. The experiments show that the approach improves performance on various tasks and has the ability to disentangle the latent factors in word embeddings. The authors also indicate that future work includes more comprehensive experimentation and theoretical analysis.

\appendix

\section{Identifiability of Word Embedding Representation}

Following~\cite{khemakhem2020variational},  we assume that embedding $\mathbf{h}$'s distribution depends on token $w$, and it is a factorial member of the exponential family with $v > 0$ sufficient statistics, see \cite{efron1975defining} for more details on exponential families.
Here $w$ is an observed variable which can be viewed as covariates.
The general form of the exponential distribution can be written~as
\begin{align}\label{eq:exp_h}
&p_{\psi}(\mathbf{h} | w) 
= \Pi_{i=1}^l \frac{Q_i(h_{i})}{Z_i(w)} \text{exp}\bigg[ \sum_{j=1}^v T_{i,j}(h_{i}) \lambda_{i,j}(w) \bigg].
\end{align}
where $Q_i$ is the base measure, $Z_i(w)$ is the normalizing constant, $T_{i,j}$ are the component of the sufficient statistic and $\lambda_{i,j}$ the corresponding parameters, depending on the variable $w$.
Data sequence $\mathbf{s}$ is generated with some complex and deterministic function from $h$ as in:
\begin{align}\label{eq:xt_gen}
\tilde{\mathbf{s}} = \mathbf{f}(\mathbf{h}, \epsilon)
\end{align}
Let $\mathbf{T} =[\mathbf{T}_1, ..., \mathbf{T}_l] $, and $\mathbf{\lambda} =[\mathbf{\lambda}_1, ..., \mathbf{\lambda}_l]$.
 We define the domain of $\mathbf{h}$ as $\mathcal{H}=\mathcal{H}_1 \times ... \times \mathcal{H}_l$.

The parameter set $\widehat{\Theta} = \{\widehat{\mathbf{\theta}} :=(\widehat{\mathbf{T}}, \widehat{\mathbf{\lambda}}, \widehat{\mathbf{f} }) \}$
 represents the model learned by a piratical algorithm. Let $\widehat{\mathbf{h}}$ be one word's latent variable recovered by the algorithm regarding $\mathbf{h}$.
In the limit of infinite data and algorithm convergence, we establish the following theoretical result regarding the identifiability of the sufficient statistics $\mathbf{T}$ in our model~(\ref{eq:exp_h}).

\textbf{Theorem 1.} {\it
Assume that the observed data is distributed according to the model given by~(\ref{eq:exp_h}) and~(\ref{eq:xt_gen}).
Let the following assumptions hold,

(a) The sufficient statistics $T_{ij}(h)$ are differentiable almost everywhere and their derivatives $ \partial T_{i,j}/\partial_h$ are nonzero almost surely for all $h\in \mathcal{H}_i$, $1\leqslant i \leqslant l$ and $1 \leqslant j  \leqslant v$.

(b) There exist $(lv+1)$ distinct conditions $w_{0}$, ..., $w_{lv}$  such that the matrix
\begin{equation*}
\mathbf{L} = [\mathbf{\lambda}(w_{1}) - \mathbf{\lambda}(w_{0}), ..., \mathbf{\lambda}(w_{lv}) - \mathbf{\lambda}(w_{0}) ]
\end{equation*}
of size $lv \times lv$ is invertible.

Then the model parameters
$\mathbf{T}(\mathbf{h}) = \mathbf{A}\widehat{\mathbf{T}}(\widehat{\mathbf{h}}) + \mathbf{c}.$ Here $\mathbf{A}$ is an $lv \times lv$ invertible matrix and $\mathbf{c}$ is a vector of size $lv$.
}

\begin{proof}
The conditional probabilities of $p_{\mathbf{T}, \mathbf{\lambda}, \mathbf{f} }\big(\tilde{\mathbf{s}} | w\big)$ and $p_{\widehat{\mathbf{T}}, \widehat{\mathbf{\lambda}}, \widehat{\mathbf{f}} }\big(\tilde{\mathbf{s}} | w\big)$ are assumed to be the same in the limit of infinite data.
By expanding the probability density functions with the correct change of variable, we have
\begin{align} \notag
& \log p_{\mathbf{T}, \mathbf{\lambda}}(\mathbf{h}| w) + \log \big| \det \mathbf{J}_{\mathbf{f}}(\tilde{\mathbf{s}}) \big|\\
=& \log p_{\widehat{\mathbf{T}}, \widehat{\mathbf{\lambda}}}(\widehat{\mathbf{h}}| w) + \log \big| \det \mathbf{J}_{\widehat{\mathbf{f}}}(\tilde{\mathbf{s}}) \big|.
\end{align}
Let $w_{0},...,w_{lv}$ be from condition (b). We can subtract this expression of $w_{0}$ from some  $w_{u}$. The Jacobian terms will be removed since they do not depend  $w$,
\begin{align} \label{eq:u_diff}
&\log p_{\mathbf{h}}(\mathbf{h}|w_{u}) - \log p_{\mathbf{h}}(\mathbf{h}|w_{0}) \\ \notag
=& \log p_{\widehat{\mathbf{h}}}(\widehat{\mathbf{h}}|w_{u}) - \log p_{\widehat{\mathbf{h}}}(\widehat{\mathbf{h}}|w_{0}) .
\end{align}
Both conditional distributions in~\eqref{eq:u_diff} belong to the exponential family.
Eq.~(\ref{eq:u_diff}) thus reads
\begin{align} \notag
&\sum_{i=1}^l \bigg[\log \frac{Z_i(w_{0})}{Z_i(w_{u})} + \sum_{j=1}^v T_{i,j}(\mathbf{h})\big(\lambda_{i,j}(w_{u}) \\ \notag
&- \lambda_{i,j}(w_{0})\big) \bigg] = \sum_{i=1}^d \bigg[\log \frac{\widehat{Z}_i(w_{0})}{\widehat{Z}_i(w_{u})} \\ \notag
&+ \sum_{j=1}^m \widehat{T}_{i,j}(\widehat{\mathbf{h}})\big(\widehat{\lambda}_{i,j}(w_{u})- \widehat{\lambda}_{i,j}(w_{0})\big) \bigg].
\end{align}
Here the base measures $Q_i$'s are canceled out.
Let $\bar{\mathbf{\lambda}}(w) = \mathbf{\lambda}(w_u)-\mathbf{\lambda}(w_{0})$.
The above equation can be expressed as
\begin{align} \notag
& \langle \mathbf{T}(\mathbf{h}), \bar{\mathbf{\lambda}}	\rangle + \sum_i \log \frac{Z_i(w_{0})}{Z_i(w_{u})}
=\langle \widehat{\mathbf{T}}(\widehat{\mathbf{h}}), \bar{\widehat{\mathbf{\lambda}}}	\rangle \\ \notag
&+ \sum_i\log \frac{\widehat{Z}_i(w_{0})}{\widehat{Z}_i(w_{u})}, \ \ \forall \ u, 1 \leqslant u \leqslant lv .
\end{align}
Combine $lv$ equations together and we can rewrite them in matrix equation form as following
\begin{align} \notag
\mathbf{L}^{\top}\mathbf{T}(\mathbf{h}) = \widehat{\mathbf{L}}^{\top}\widehat{\mathbf{T}}(\widehat{\mathbf{h}}) + \mathbf{b}.
\end{align}
Here $b_u=\sum_{i=1}^{l}\log \frac{\widehat{Z}_i(w_{0}) Z_i(w_{u}) }{\widehat{Z}_i(w_{u}) Z_i(w_{0}) }$. We can multiply $\mathbf{L}^{\top}$'s inverse with both sized of the equation,
\begin{align}\label{eq:A_sim}
\mathbf{T}(\mathbf{h}) = \mathbf{A}\widehat{\mathbf{T}}(\widehat{\mathbf{h}}) + \mathbf{c}.
\end{align}
Here $\mathbf{A} = \mathbf{L}^{-1\top} \widehat{\mathbf{L}}^{\top} $, and $\mathbf{c} = \mathbf{L}^{-1\top} \mathbf{b}$.
By Lemma 1 from~\cite{khemakhem2020variational}, there exist $v$ distinct values $h^{i}_{1}$ to $h^{i}_{v}$ such that $\big[ \frac{d T_i}{ d h^{i}}(h^{i}_{1}), ...,  \frac{d T_i}{ d h^{i}}(h^{i}_{v}) \big]$ are linearly independent in $\mathbb{R}^v$, for all $1\leqslant i \leqslant l$.
Define $v$ vectors $\mathbf{h}_{u}= [h^{1}_u, ..., h^{l}_u]$ from points given by this lemma.
We obtain the  Jacobian matrix
$$\mathbf{Q}= [\mathbf{J}_{\mathbf{T}}(\mathbf{h}_1), ..., \mathbf{J}_{\mathbf{T}}(\mathbf{h}_v)] \, ,$$
where each entry is the Jacobian of size $lv \times l$ from the derivative of Eq.~(\ref{eq:A_sim}) regarding the $m$ vectors $\{\mathbf{h}_j\}_{j=1}^v$.
Hence $\mathbf{Q}$ is an $lv \times lv$ invertible by the lemma and the fact that each component of $\mathbf{T}$ is univariate. 
We can construct a corresponding matrix $\widehat{\mathbf{Q}}$ with the Jacobian of $\widehat{\mathbf{T}}(\widehat{\mathbf{f}}^{-1}\circ \mathbf{f}^{-1}(\mathbf{h}))$ computed at the same points and get
\begin{align} \notag
\mathbf{Q} = \mathbf{A}\widehat{\mathbf{Q}} \,.
\end{align}
Here $\widehat{\mathbf{Q}}$ and $\mathbf{A}$ are both full rank as $\mathbf{Q}$ is full rank.
\end{proof}

{\small
\bibliographystyle{plain}
\bibliography{ref}
}

\end{document}